\documentclass{article}
\usepackage[dblblindworkshop,final]{neurips_2026}
\workshoptitle{Agentic AI Benchmarks and Applications for Enterprise Tasks}

\usepackage[utf8]{inputenc}
\usepackage[T1]{fontenc}
\usepackage{hyperref}
\usepackage{url}
\usepackage{booktabs}
\usepackage{amsmath}
\usepackage{amssymb}
\usepackage{graphicx}
\usepackage{microtype}
\usepackage{xcolor}
\usepackage{enumitem}
\usepackage{wrapfig}
\usepackage{float}
\DeclareMathOperator*{\argmin}{arg\,min}

\newcommand{\ufs}{\mathrm{UFS}}

\title{UserProxyBench: Evaluating LLM User Simulators for Agent Benchmarks and Training}
\author{%
  Ashish Jain\\
  \texttt{ashish@sarvam.ai}
  \And
  Armaan Sandhu\\
  \texttt{apsandhu@umass.edu}
}

\begin{document}
\maketitle

\begin{abstract}
Interactive agent benchmarks and multi-turn reinforcement learning increasingly place a second language model in the role of the user. This simulated user controls what information the agent receives and when, yet current benchmarks score only the agent and do not directly measure whether the user correctly executed its assigned role. We introduce UserProxyBench, an evaluation layer over the \(\tau\)-bench family, and the User Fidelity Score (UFS), which measures adherence to the benchmark's private user instructions using task-grounded rubric criteria scored independently of agent success. Holding the agent fixed at GPT-5.5 and varying only the user proxy across 375 enterprise tasks changes mean task reward by 15.2 points, while 24.4\% of successful episodes contain a user-specification violation. The dominant failure is premature disclosure: users provide information before it is requested. This behavior has little effect on task reward, yet among successful episodes it causes the agent to make 1.06 fewer tool calls on average, changing the interaction being evaluated while preserving the reward. Finally, across seven proxies we identify an empirical cost--fidelity frontier, enabling practitioners to select the least expensive simulator that satisfies a required fidelity level.
\end{abstract}

\section{Introduction}
Interactive evaluation has made the ``user'' part of the benchmark executable. In $\tau$-bench, an LLM user converses with an agent that follows enterprise policy and uses tools \citep{yao2024taubench}; $\tau^2$-bench extends this to dual-control settings where the user can also act on a shared environment \citep{barres2025tau2}. The same pattern is entering training: MUA-RL places an LLM-simulated user inside the reinforcement-learning loop \citep{zhao2025muarl}, while broader agent-RL systems optimize over long interactive trajectories \citep{wang2025ragen,luo2025agentlightning}.

This creates an experimental dependency that is easy to overlook. The user proxy controls information timing, observations, and sometimes user-side actions. If it leaks information, fabricates state, or stops early, the agent is no longer solving the intended interaction even when task reward is unchanged. For example, one Telecom proxy opens with ``I'm John Smith, by the way. My number is 555-123-2002.'' before the agent asks for either field, removing the information-gathering step the task was designed to exercise.

Recent work shows that LLM users differ from real humans and that simulator choice can change measured assistant performance \citep{dou2025simulatorarena,seshadri2026lost,zhou2026sim2real}. Purpose-built user models improve human-likeness over assistant LMs prompted to play users \citep{naous2026userlm,wu2026humanlm}. We ask a complementary question: \emph{given the benchmark's own stated user specification, did the proxy execute that role correctly?} We call this \textbf{functional fidelity}. It measures contract adherence, not human realism.

We make three contributions. First, we introduce \textbf{UserProxyBench}, an evaluation layer over $\tau$ tasks that scores the user independently of the agent. Second, we show that task reward does not subsume user fidelity: nearly one quarter of successful trajectories contain a user-contract violation, and the most common failure is almost invisible to agent reward. Third, we treat user-proxy choice as an explicit operational trade-off, using UFS as a reliability constraint under which user-side inference cost can be minimized for a given evaluation or training pipeline.

\section{UserProxyBench}
\label{sec:userproxybench}
\paragraph{User Fidelity Score.}
Each task provides a private user blueprint (persona, known and unknown information, goal, and task-specific instructions), standing simulator guidelines, and, in some domains, user tools. Let $C_i=\{c_{i1},\ldots,c_{im}\}$ be the applicable criteria for episode $i$. We define
\begin{equation}
 f_i=\prod_{j=1}^{m}\mathbf{1}\!\left\{c_{ij}\ \text{passes}\right\}, \qquad
 \ufs=\frac{1}{N}\sum_{i=1}^{N}f_i .
\end{equation}
UFS is therefore the fraction of episodes in which the proxy satisfies every applicable criterion in its own contract. A separate judge-free user-action probe is reported for Telecom, where gold user write actions exist, but is not folded into UFS because the other domains do not expose the same signal.

\paragraph{Task-grounded rubrics.}
Claude Opus 4.8 writes 4--8 atomic criteria per task from the private blueprint, the exact simulator guidelines, the user tool surface, and one sample interaction. The sample is used only to identify situations that can arise; the blueprint and guidelines remain the authority for correctness. Each criterion must cite a verbatim grounding span, be falsifiable on a trajectory, and concern only the user. An independent verifier (GPT-5.6-Sol, distinct from the generator) reviews each task-level rubric set for grounding, achievability, redundancy, and scope before scoring. Claude Opus 4.8 then grades candidate trajectories without seeing proxy identity; generator and grader are the same model, a limitation we return to in Section~\ref{sec:discussion}. The final set contains 2,141 criteria: 634 Telecom, 296 Airline, 656 Retail, and 555 Banking (5.7 per task), spanning four families: \emph{groundedness}, \emph{premature disclosure}, \emph{goal deviation}, and \emph{missed information}. This follows the broader use of instance-specific rubrics for open-ended evaluation and reward construction \citep{arora2025healthbench,gunjal2026rar}, but targets the \emph{user-side environment component} rather than the assistant response.

The dominant family derives from an explicit instruction shown to every proxy: \emph{``Disclose information progressively. Wait for the agent to ask for specific information before providing it.''} Similar explicit rules prohibit inventing unavailable information, require grounding tool results, and require continuing until the task goal is satisfied. We therefore score stated contract compliance rather than an implicit style preference.

\paragraph{Setup.}
We evaluate seven reported proxies on the full base split of four enterprise domains: Telecom (114 tasks), Airline (50), Retail (114), and Banking Knowledge (97), for 375 tasks total. GPT-5.5 is frozen as the agent; only the user proxy changes. The reported set contains three open-weight and four hosted models. Each task has one rollout per proxy. We denote the benchmark's agent-side task reward by $R_{\tau}$ to distinguish it from the $\tau$ benchmark family. Hosted costs use billed user-side spend; for self-hosted models, measured input/output tokens are converted using public serverless prices accessed in August 2026 \citep{together2026pricing}.

\section{Results}
\subsection{Fidelity-constrained proxy selection}
\label{sec:selection}
\begin{figure}[t]
\centering
\includegraphics[width=.72\linewidth]{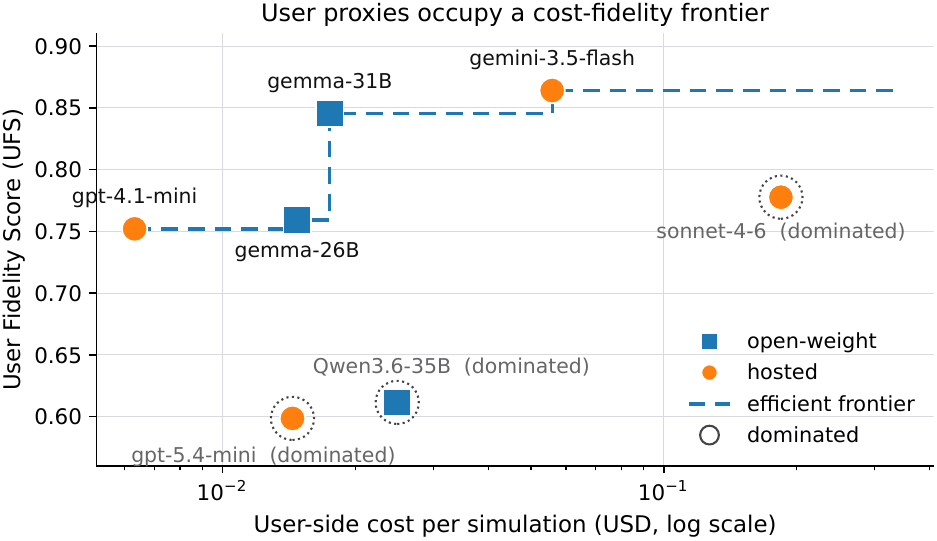}
\caption{\textbf{Cost--fidelity frontier.} Each point is one user proxy; the agent is frozen at GPT-5.5. The dashed line is the efficient frontier (no proxy is both cheaper and higher-UFS). Ringed points are strictly dominated. Because the axis is log-cost, the least-expensive proxy meeting a required UFS $q$ changes with $q$: Gemma-31B at $q=.84$, Gemini-3.5-Flash at $q=.86$.}
\label{fig:cost}
\end{figure}

Once fidelity is measurable, the operational question is which simulator is the least expensive one that meets the pipeline's reliability requirement. For a required fidelity $q$,
\begin{equation}
U^*(q)=\argmin_U C(U) \quad \text{s.t.} \quad \ufs(U)\ge q,
\end{equation}
where $C(U)$ is user-side inference cost. Figure~\ref{fig:cost} shows the frontier. At $q=.84$, \textbf{Gemma-31B is the lowest-cost qualifying proxy}: UFS .845 at about \$0.0175 per simulation. Gemini-3.5-Flash gains 1.9 UFS points but costs 3.2$\times$ more; raising the requirement to $q=.86$ moves the operating point to Gemini, while at $q=.75$ GPT-4.1-mini is cheaper. Three tested proxies are strictly dominated because another model is both cheaper and more faithful. At 64 rollouts over 1,000 prompts, Gemma-31B's user-side inference is about \$1.1k per epoch versus \$3.6k for Gemini-3.5-Flash.

\subsection{The user proxy changes the benchmark result}
\begin{figure}[t]
\centering
\includegraphics[width=.68\linewidth]{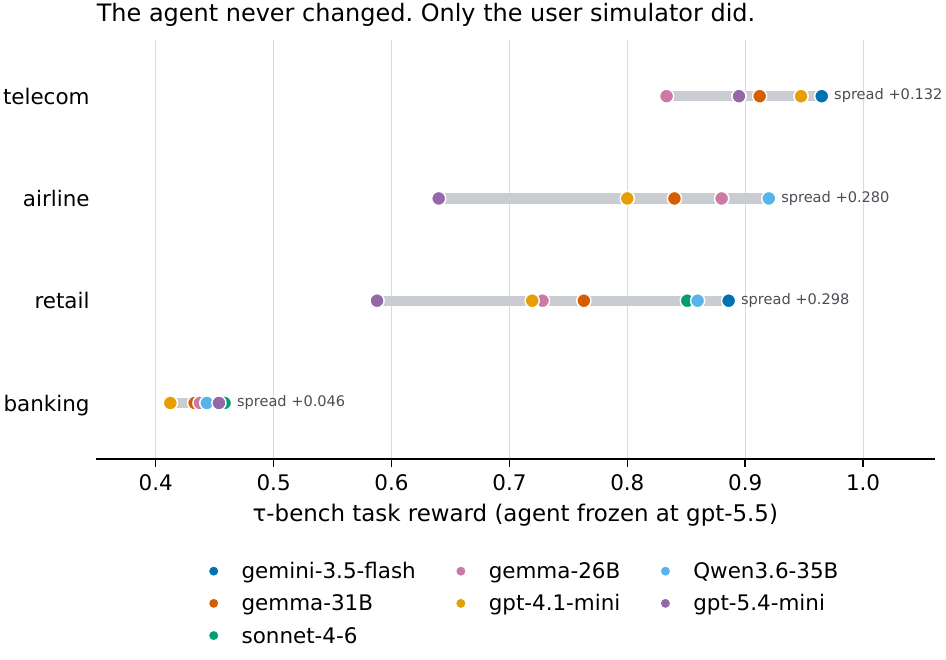}
\caption{\textbf{Only the user simulator changed.} Each point is one of seven user proxies; the grey bar spans the $R_{\tau}$ range induced for the same frozen GPT-5.5 agent within a domain.}
\label{fig:spread}
\end{figure}

Nothing about the evaluated agent changes across proxy conditions, yet mean $R_{\tau}$ ranges from .644 to .796. Excluding Banking, whose absolute reward is uniformly low, per-domain spread ranges from .132 to .298, reaching .280 on Airline and .298 on Retail. Figure~\ref{fig:spread} makes the experimental control explicit: every point on a row is the same GPT-5.5 agent on the same domain, with only the simulated user changed.

Across all 2{,}618 scored trajectories (of 2{,}625: two runs hit an infrastructure error, and five were unscorable because no criterion applied---Appendix~B), 24.4\% of agent-successful episodes fail UFS; on Telecom, 39.7\% of all trajectories are $R_{\tau}$ pass / UFS fail.

\subsection{Why task reward cannot screen user proxies}
\begin{figure}[t]
\centering
\includegraphics[width=.70\linewidth]{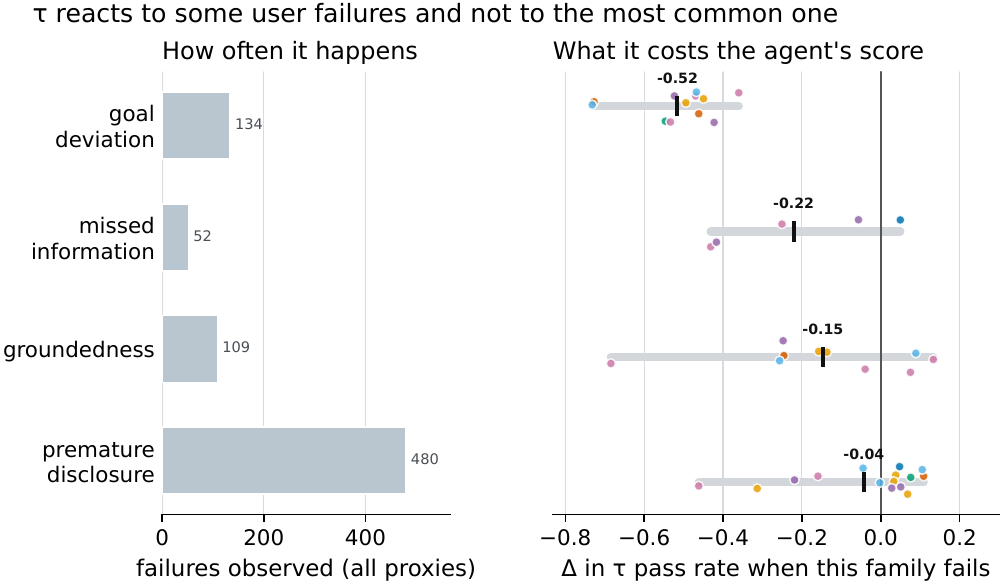}
\caption{\textbf{$R_{\tau}$ is nearly blind to the dominant failure.} Left: how often each failure family occurs across all proxies. Right: the change in agent pass rate when the family is present versus absent, one dot per proxy--domain cell with at least five episodes on both sides (up to 28 cells). Goal deviation costs the agent $-.52$; premature disclosure, the most common failure, costs $-.04$.}
\label{fig:mechanism}
\end{figure}

Within each proxy--domain cell, we compare agent pass rate when a failure family occurs versus when it does not (at least five episodes per side). Goal deviation is uncommon (134 failures) and reduces pass rate by .515 on average. Premature disclosure is the dominant failure (480), yet changes pass rate by only $-.043$ on average; its median effect is \textbf{positive} ($+.033$) and the sign is inconsistent across cells. Figure~\ref{fig:mechanism} shows both the frequency of each family and its effect on agent reward: the most common failure is the one $R_{\tau}$ is least able to detect.

This is visible even for capable models. On Telecom, Sonnet-4-6 and GPT-4.1-mini give the frozen agent the same $R_{\tau}=.947$, but their UFS scores are .474 and .737. The gap is not only disclosure: in another successful Telecom episode, Gemma-26B says it is turning off Airplane Mode although no corresponding user tool call occurs and the device remains in Airplane Mode (Appendix~C).

Removing disclosure criteria substantially reorders models: Qwen3.6-35B moves from sixth to third, Gemma-26B from fourth to last, and disclosure-only versus non-disclosure UFS are almost uncorrelated ($\rho=.071$). UFS is therefore best read together with its failure profile.

\subsection{Implications for multi-turn RL}
For training, the issue is not only whether the final reward changes; it is whether the \emph{trajectory being reinforced} changes. Restricting to successful episodes ($R_{\tau}=1$), over-disclosing users make the frozen agent issue 1.06 fewer tool calls and ask 0.26 fewer questions, consistently across all seven proxies. The agent receives identical task reward despite doing less information gathering.

Across the same 374 tasks completed by all seven proxies, mean agent turns correlate with UFS at $r=.955$, against $r=.594$ for $R_{\tau}$ (Appendix~\ref{app:rlfig}); proxy choice changes mean turns by 26\% and questions by 25\%. In multi-turn RL, fidelity determines whether rollouts instantiate the intended interaction, while cost determines how many such rollouts fit the budget. Hence the operating rule in Section~\ref{sec:selection}: choose the lowest-cost proxy that meets the required fidelity.

\section{Discussion}
\label{sec:discussion}
UserProxyBench measures contract fidelity, not human realism; the two are complementary. UFS is LLM-judged by the rubric generator, and our manual audit (ten tasks plus 50 criteria/judgments) is single-rater. A judge-free Telecom probe finds 54 missed gold writes among 658 trajectories; of these, 23 pass every applicable criterion and 20 pass a directly relevant one. Limitations include one rollout per task, time-dependent serving prices, a pinned Banking version, and no measurement of downstream trained-policy effects.

\appendix

\section{Full proxy results}
\begin{table}[H]
\centering
\small
\setlength{\tabcolsep}{3.1pt}
\begin{tabular}{lrrrrrr}
\toprule
Proxy & Hosting & Cost/sim & Agent $R_{\tau}$ & Mean UFS & Prompt tok & Completion tok \\
\midrule
Sonnet-4-6$^{\dagger}$ & hosted & .1841 & .784 & .778 & 57,888 & 665 \\
Gemini-3.5-Flash & hosted & .0558 & .796 & .864 & 53,111 & 2,026 \\
Qwen3.6-35B$^{\dagger}$ & open & .0249 & .779 & .612 & 47,590 & 7,524 \\
Gemma-31B & open & .0175 & .737 & .845 & 43,696 & 461 \\
Gemma-26B & open & .0147 & .720 & .759 & 36,411 & 562 \\
GPT-5.4-mini$^{\dagger}$ & hosted & .0144 & .644 & .599 & 26,050 & 1,587 \\
GPT-4.1-mini & hosted & .0063 & .720 & .752 & 29,574 & 449 \\
\bottomrule
\end{tabular}
\caption{Aggregate proxy results, sorted by user-side cost. Agent $R_{\tau}$ and Mean UFS are unweighted means across the four domains for the frozen GPT-5.5 agent under each proxy. $^{\dagger}$Strictly dominated on the cost--UFS plane.}
\label{tab:aggregate}
\end{table}

\section{Per-domain $R_{\tau}$ and UFS}
\begin{table}[H]
\centering
\scriptsize
\setlength{\tabcolsep}{2.4pt}
\begin{tabular}{lrrrrrrrr}
\toprule
& \multicolumn{2}{c}{Telecom} & \multicolumn{2}{c}{Airline} & \multicolumn{2}{c}{Retail} & \multicolumn{2}{c}{Banking} \\
Proxy & $R_{\tau}$ & UFS & $R_{\tau}$ & UFS & $R_{\tau}$ & UFS & $R_{\tau}$ & UFS \\
\midrule
Sonnet-4-6 & .947 & .474 & .880 & .840 & .851 & .912$^{\dagger}$ & .458 & .885$^{*}$ \\
Gemini-3.5-Flash & .965 & .711 & .880 & .940 & .886 & .929$^{\dagger}$ & .454 & .876 \\
Qwen3.6-35B & .895 & .254 & .920 & .680 & .860 & .646$^{\dagger}$ & .443 & .866 \\
Gemma-31B & .912 & .781 & .840 & .880 & .763 & .886 & .433 & .835 \\
Gemma-26B & .833 & .588 & .880 & .800 & .728 & .868 & .438 & .781$^{*}$ \\
GPT-5.4-mini & .895 & .325 & .640 & .580 & .588 & .717$^{\dagger}$ & .454 & .773 \\
GPT-4.1-mini & .947 & .737 & .800 & .740 & .719 & .779$^{\dagger}$ & .412 & .753 \\
\bottomrule
\end{tabular}
\caption{Per-domain results. $n=114/50/114/97$ for Telecom/Airline/Retail/Banking, with two exceptions: Banking $n=96$ in the two cells marked $^{*}$ (infrastructure error on one task), and Retail $n=113$ for five proxies, marked $^{\dagger}$, where one task yielded no applicable criteria and UFS is undefined. 2,618 of 2,625 trajectories are scored.}
\label{tab:perdomain}
\end{table}

\section{Failure taxonomy and examples}
The taxonomy defines 19 modes under four families; 18 are exercised by the 2,141 retained criteria. \texttt{groundedness.self\_contradiction} receives zero criteria on this task set. Premature disclosure contains \texttt{volunteers\_unasked\_info}, \texttt{preempts\_diagnosis}, and \texttt{dumps\_blueprint}; groundedness covers unsupported personal/device state, unexecuted-action claims, contradictions, and invented unknown information; goal deviation covers premature stopping, accepting forbidden or insufficient outcomes, and abandoning the goal; missed information covers withholding requested information and skipping requested user actions.

\paragraph{Premature disclosure (Qwen3.6-35B, Telecom, $R_{\tau}=1$).}
\begin{quote}\small
``I'm John Smith, by the way. My number is 555-123-2002.''
\end{quote}
This is the opening user turn, before the agent requests identifying information.

\paragraph{Unexecuted action claim (Gemma-26B, Telecom, $R_{\tau}=1$).}
\begin{quote}\small
``Okay, I'll try that. Let me see... I'm turning off this Airplane Mode now. Just a moment... okay, I've waited a bit.''
\end{quote}
No \texttt{toggle\_airplane\_mode} call occurred, and subsequent device state still showed Airplane Mode on.

\paragraph{Fabricated device state (Gemma-26B, Telecom).}
\begin{quote}\small
``Let me try to send that picture one more time... Oh! It worked! The little bar finished moving and it says it was sent!''
\end{quote}
No tool result established MMS success; the last relevant check had returned failure.

\section{Rubric construction and judge-free probe}
Claude Opus 4.8 generates criteria from the private user blueprint, simulator guidelines, user tool surface, and one sample conversation; the prompt explicitly states that the sample is not an oracle. Criteria must cite an exact grounding span and are batch-reviewed by GPT-5.6-Sol under nine checks covering grounding, falsifiability, scope, user-attribution, achievability, taxonomy fit, and redundancy. Claude Opus 4.8 then grades candidate trajectories one criterion at a time without the proxy identity.

A judge-free Telecom probe compares actual user tool calls against gold user write assertions on 658 scorable trajectories. Fifty-four trajectories miss a gold write; in 23 of these every applicable semantic criterion still passes, and in 20 a directly relevant criterion is applied and still passes. Adding this mechanical signal would change Telecom UFS by .009--.070 per proxy and leaves the ranking unchanged, so we report it as an independent diagnostic rather than changing the cross-domain UFS definition.

\section{Agent effort versus proxy metrics}
\label{app:rlfig}
\begin{figure}[t]
\centering
\includegraphics[width=.86\linewidth]{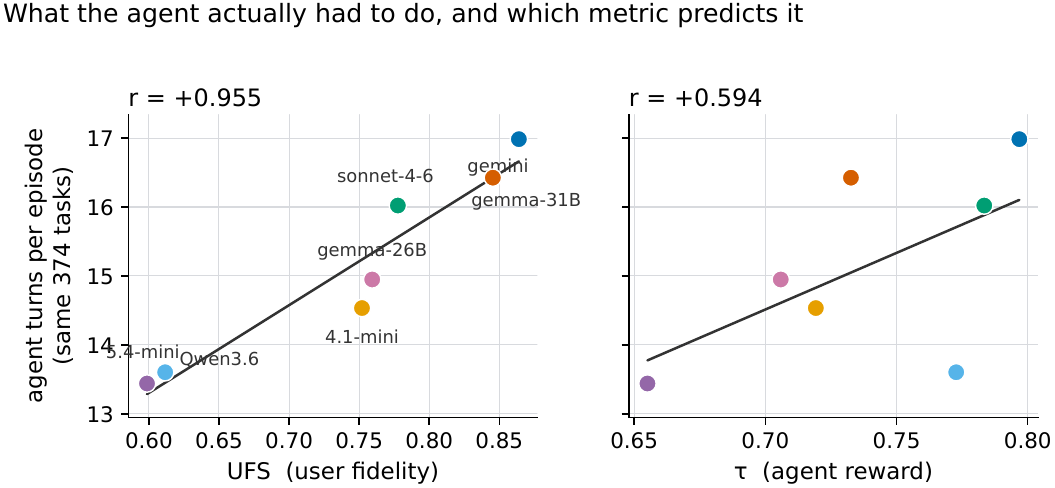}
\caption{Mean agent turns over the 374 tasks completed by all seven proxies, against UFS (left) and agent $R_{\tau}$ (right). Points are directly labeled by proxy.}
\label{fig:rl-signal}
\end{figure}

Figure~\ref{fig:rl-signal} compares UFS and agent task reward as predictors of the interaction induced by each proxy. Across the 374 tasks completed by all seven proxies, mean agent turns correlate with UFS at $r=.955$, compared with $r=.594$ for $R_{\tau}$.

\section{Limitations and next validation}
\label{app:limitations}
Our current manual audit consists of ten tasks read end-to-end plus 50 further final criteria/judgments reviewed by one author. It changed the pipeline twice but is not blinded, stratified, or multi-rater. The most important next validation is therefore a blinded human study of criterion validity and judge decisions, with agreement statistics. Repeated rollouts on Telecom, a second frozen agent, and a small RL experiment comparing policies trained against high- and low-UFS proxies would separately test stochasticity, interaction-partner dependence, and downstream training effects.

\section{Experimental configuration and assets}
\label{app:config}
All seven reported user proxies use temperature 1.0. The frozen GPT-5.5 agent uses high reasoning effort. GPT-5.4-mini also uses high reasoning effort; GPT-4.1-mini has no thinking mode; Gemini-3.5-Flash uses its default thinking behavior; Qwen3.6-35B-A3B is served locally with thinking enabled; Gemma-4-31B-it and Gemma-4-26B-A4B-it are served locally with thinking disabled; Sonnet-4-6 was run without extended thinking because the trajectory runner did not forward the provider-specific parameter. Local Gemma/Qwen inference used tensor parallelism over two workers (Gemma-4-26B-A4B-it and Qwen also used expert parallelism).

Rubric criteria were generated by Claude Opus 4.8 (temperature 1.0), independently verified by GPT-5.6-Sol, and graded by Claude Opus 4.8. Generator and judge are therefore the same model. We exclude Opus-4-8 and GPT-5.5 from the reported proxy set because their conversations or outputs participate in rubric construction; GPT-5.5 remains the frozen \emph{agent} in all runs. The benchmark is built on sierra-research/tau2-bench v1.0.0 at commit \texttt{8ebb749} (MIT license), with two local engineering patches for judge routing and context-overflow retrieval. Numerical tables and figures are regenerated deterministically from graded trajectory files.

\section{Broader impact}
\label{app:impact}
More reliable user proxies can make interactive benchmarks and RL environments easier to audit and cheaper to operate. The main risk is over-interpreting contract fidelity as human realism: a high-UFS simulator may still underrepresent real users, dialects, accessibility needs, or adversarial behavior. UFS should therefore complement, not replace, evaluations with real users and population-sensitive simulation metrics. Cost-frontier results are time-dependent and should not be treated as long-lived commercial recommendations.
\end{document}